\documentclass[conference]{IEEEtran}
\IEEEoverridecommandlockouts
\usepackage{cite}
\usepackage{amsmath,amssymb,amsfonts}
\usepackage{algorithmic}
\usepackage{graphicx}
\usepackage{textcomp}
\usepackage{xcolor}
\usepackage{hyperref}

\usepackage{amsmath}
\usepackage{amsfonts}
\newtheorem{theorem}{Theorem}
\newtheorem{lemma}{Lemma}

\usepackage{multirow}
\usepackage{mathrsfs}
\usepackage{graphics}
\usepackage{subfig}
\usepackage{wrapfig}
\usepackage{epstopdf}
\usepackage{pdfpages}
\usepackage{diagbox}
\usepackage{ulem}
\usepackage{CJKulem}
\usepackage{MnSymbol}

\usepackage{tabularx}
\usepackage{diagbox}
\usepackage{makecell}

\usepackage{float}

\usepackage{xcolor,color}
\usepackage{soul}
\soulregister\cite7 
\soulregister\ref7 

\def\BibTeX{{\rm B\kern-.05em{\sc i\kern-.025em b}\kern-.08em
    T\kern-.1667em\lower.7ex\hbox{E}\kern-.125emX}}
\begin{document}

\title{Enhancing the Rationale-Input Alignment for Self-explaining Rationalization\\

\thanks{
This work is supported by National Natural Science Foundation of China under grants 62376103, 62302184, 62206102, and Science and Technology Support Program of Hubei Province under grant 2022BAA046.}
}

    \author{\IEEEauthorblockN{1\textsuperscript{st} Wei Liu}
\IEEEauthorblockA{\textit{School of Computer Science and } \\
\textit{Technology, Huazhong University}\\
\textit{ of Science and Technology}\\
Wuhan, China \\
idc\_lw@hust.edu.cn}
\and
\IEEEauthorblockN{2\textsuperscript{nd} Haozhao Wang\thanks{Haozhao Wang is the corresponding author.}
}
\IEEEauthorblockA{\textit{School of Computer Science and } \\
\textit{Technology, Huazhong University}\\
\textit{ of Science and Technology}\\
Wuhan, China \\
hz\_wang@hust.edu.cn}
\and
\IEEEauthorblockN{3\textsuperscript{rd} Jun Wang}
\IEEEauthorblockA{\textit{iWudao Tech} \\
Nanjing, China \\
jwang@iwudao.tech}
\and
\IEEEauthorblockN{4\textsuperscript{th} Zhiying Deng}
\IEEEauthorblockA{\textit{School of Computer Science and } \\
\textit{Technology, Huazhong University}\\
\textit{ of Science and Technology}\\
Wuhan, China \\
dengzhiyingdd@hust.edu.cn}
\and
\IEEEauthorblockN{5\textsuperscript{th} Yuankai Zhang}
\IEEEauthorblockA{\textit{School of Computer Science and } \\
\textit{Technology, Huazhong University}\\
\textit{ of Science and Technology}\\
Wuhan, China \\
yuankai\_zhang@hust.edu.cn}
\and
\IEEEauthorblockN{6\textsuperscript{th} Cheng Wang}
\IEEEauthorblockA{\textit{School of Cyber Science and } \\
\textit{Engineering, Huazhong University}\\
\textit{ of Science and Technology}\\
Wuhan, China \\
wangcheng20@hust.edu.cn}
\and
\IEEEauthorblockN{7\textsuperscript{th} Ruixuan Li}
\IEEEauthorblockA{\textit{School of Computer Science and } \\
\textit{Technology, Huazhong University}\\
\textit{ of Science and Technology}\\
Wuhan, China \\
rxli@hust.edu.cn}
}

\normalem
\maketitle

\begin{abstract}
Rationalization empowers deep learning models with self-explaining capabilities through a cooperative game, where a generator selects a semantically consistent subset of the input as a rationale, and a subsequent predictor makes predictions based on the selected rationale.
In this paper, we discover that rationalization is prone to a problem named \emph{rationale shift}, which arises from the algorithmic bias of the cooperative game. Rationale shift refers to a situation where the semantics of the selected rationale may deviate from the original input, but the predictor still produces accurate predictions based on the deviation, resulting in a compromised generator with misleading feedback. 
    To address this issue, we first demonstrate the importance of the alignment between the rationale and the full input through both empirical observations and theoretical analysis. Subsequently, we introduce a novel approach called DAR (\textbf{D}iscriminatively \textbf{A}ligned \textbf{R}ationalization), which utilizes an auxiliary module pretrained on the full input to discriminatively align the selected rationale and the original input. We theoretically illustrate how DAR accomplishes the desired alignment, thereby overcoming the rationale shift problem. {The experiments on two widely used real-world benchmarks show that the proposed method significantly improves the explanation quality (measured by the overlap between the model-selected explanation and the human-annotated rationale) as compared to state-of-the-art techniques.} Additionally, results on two synthetic settings further validate the effectiveness of DAR in addressing the rationale shift problem. 
\end{abstract}


\section{Introduction}
Recent successes of deep learning in automatically processing large volumes of data have raised concerns about the interpretability of these models, particularly given their increasing use in critical fields \cite{lipton2016mythos,xiang2019interpretable,miller2019explanation,sun2021interpreting}. Exploring  \emph{e\textbf{X}plainable \textbf{A}rtificial \textbf{I}ntelligence} (XAI) techniques can help to address many important problems existed in present deep learning models. For instance, XAI techniques can aid in detecting model discrimination (fairness) \cite{sigmod-debug}, identifying backdoor attacks (security) \cite{li2022backdoor}, and revealing potential failure cases (robustness) \cite{danqi}, among others.     
Ideally, the explanation for a prediction should be both faithful (reflective of model’s actual behavior) and plausible (matching with human understanding) \cite{Unirex}. 
Although there have been various methods to generate post-hoc explanations that may appear plausible, they may not faithfully represent an agent’s decision, because the process of generating explanations is trained separately from the model's predictions \cite{lipton2016mythos}. In the context of neural networks, particularly when assisting with critical decision-making, faithfulness should be considered a necessary prerequisite prior to plausibility. This is because faithfulness determines the trustworthiness of the explanations, which is crucial for understanding and relying on the system's decisions.

\begin{figure}
    \centering
    \includegraphics[width=0.99\columnwidth]{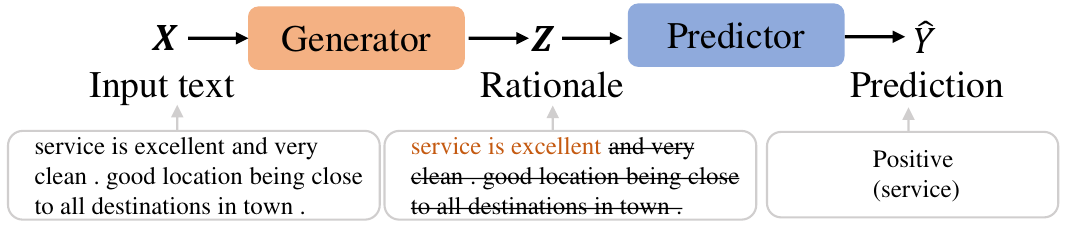}
    \caption{An example  {(binary text classification task)}  of the standard rationalization framework RNP. $X, Z, \hat{Y}$ stand for the full input, the rationale, and the prediction, respectively.}
    \label{fig:rnp}
\end{figure}

In contrast to post-hoc methods that generate  explanations for a black-box model after the prediction is made, ante-hoc self-explaining techniques generally offer greater transparency \cite{lipton2016mythos} and faithfulness \cite{interlocking,invarant} since the prediction is made based on the explanation itself.  
A model-agnostic self-explaining framework called Rationalizing Neural Predictions (RNP) has been proposed by Lei et al. \cite{emnlp/LeiBJ16}, also known as rationalization.  {RNP employs a cooperative game between a generator and a predictor, in which the generator identifies a human-interpretable subset of the input (known as the rationale) and passes it to the subsequent predictor for making predictions, as shown with an example in Fig.~\ref{fig:rnp} (an example of binary text classification). The generator and predictor are trained cooperatively to maximize the prediction accuracy. The interpretability is demonstrated through validating whether the generated rationale aligns with expectation\cite{interlocking}}. A unique advantage of RNP-based rationalization is certification of exclusion, i.e, any unselected part of the input is guaranteed to have no contribution to the prediction \cite{interlocking}. Another advantage is training under explanation (i.e., the predictor can only see the rationale during training), which ensures the maintenance of faithfulness, allowing the focus to be solely on plausibility. Thanks to these advantages, RNP and its variants have become one of the mainstream to facilitate the interpretability of NLP models \cite{dmr,car,rethinking,invarant,interlocking,liufr}. 
Notably, while RNP was originally introduced in the field of natural language processing (NLP) and its improvement schemes are primarily validated using text data, it exhibits remarkable versatility and can be easily adapted for other domains with minor adjustments. For instance, it can be employed to explain image classifiers \cite{GDM} and graph neural networks \cite{pgexplainer}.

\begin{figure}
    \centering
    \includegraphics[width=0.99\columnwidth]{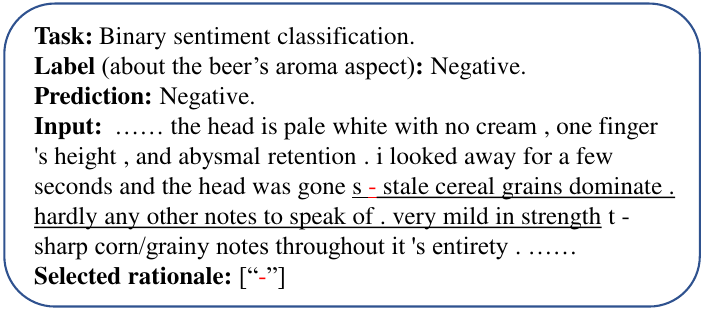}
    \caption{A cherry-picked example of RNP making the correct sentiment prediction using the uninformative rationale. 
The \underline{underlined} and \textcolor{red}{red} pieces of the text are the human-annotated and RNP-selected rationales, respectively. 
Initially, the generator may randomly select some uninformative candidates like “-” as rationales for the negative text. 
The predictor of RNP overfits to these uninformative rationales and classifies the sentiment according to whether the punctuation “-” is included in the rationale. 
Guided by such a spoiled predictor, the generator in turn tends to select these uninformative rationales.}
    \label{fig:egdegeneration}
\end{figure}

Although RNP offers a promising self-explaining framework, we identify it suffers from a serious algorithmic bias problem, namely \emph{rationale shift}, which prevents the generator from selecting rationales with good plausibility.  Specifically, the generator may add some trivial but distinguishable patterns to the selected rationales, and they will still get correct classification labels from the predictor, even if the semantics of these rationales deviate from the raw input.
Under this case, the predictor may overfit to the deviation contained in the rationales rather than to the true semantics, subsequently providing misleading feedback to the generator, encouraging the selection of these trivial patterns even more. 
A cherry-picked example of this scenario is presented in Fig. \ref{fig:egdegeneration}.
This situation arises because in a cooperative game, the primary objective of both players is to accommodate each other \cite{AAAI20sterberg}, potentially leading to collusion.

Intuitively, a good predictor should also make the right prediction when using the full input $X$, because the full input contains complete information, and generally it is expected to be more indicative to the target label than the rationale that contains only partial information (please refer to Lemma~\ref{lem:x get lower entropy than z} and the corresponding remark).
However, this is not always the case for RNP's predictor due to the occurrence of rationale shift. As a practical example of rationale shift, Fig.~\ref{fig:rationale_acc2} shows that sometimes the predictor of RNP achieves a high classification accuracy for the rationales selected by the generator but fails to classify the original full input, which indicates that the predictor also deviates due to deviations in rationales.

One special case of the problems caused by rationale shift is the degeneration\footnote{ Consider a situation where the generator has implicitly learned the
category of $X$, and selects a ``-” for all negative inputs while excluding it from all positive inputs.
In this case, the predictor only needs to determine whether the input rationale includes a ``-” or not. The example is from \cite{liufr}} problem \cite{rethinking}, where the rationales selected by the generator are solely noise, which is illustrated with an example in Fig.~\ref{fig:egdegeneration}.
Some previous efforts have been proposed to fix this special case of the rationale shift problem in rationalization, i.e., the degeneration problem. The main idea of these methods is to adopt one or more extra modules to provide calibrated information for the rationalization models.
For example, 3PLAYER~\cite{rethinking} adopts an extra predictor to squeeze informative parts from the unselected text pieces into the rationale. 
DMR~\cite{dmr} employs an extra predictor whose input is the full text and utilizes its logits as the calibrated information to the predictor. 
A2R~\cite{interlocking} leverages an implicitly extra predictor with the smoothed rationales as the input to regularize the original predictor. 
Although these methods can somewhat fix the degeneration problem of rationalization, their extra modules are required to be coordinatively trained with rationalization models from scratch and thus themselves may be affected by the deviation, which still has difficulties in solving the general case of rationale shift in practice.

To tackle this challenge, we propose a novel rationalization method architecture DAR (\textbf{D}iscriminatively \textbf{A}ligned \textbf{R}ationalization), which utilizes the \textbf{d}iscriminative \textbf{a}lignment of a pretrained predictor as the calibrated information. The architecture of DAR is shown in Fig.~\ref{fig:gr}. 
Specifically, motivated by adversarial learning methods~\cite{gan} where two distributions can be aligned, we employ a predictor pretrained with the full input as a discriminator. The discriminator acts as a third-party rationale quality inspector, ensuring alignment between the selected rationale and the raw input. Consequently, the original predictor (the blue module in Fig.~\ref{fig:rnp} and~\ref{fig:gr}) won't be disturbed by the deviation, enabling the generator to receive appropriate feedback.
On the other side, unlike existing adversarial learning methods where the discriminator is coordinatively trained with the generator using both original data and generated data, the discriminative predictor of DAR is pretrained only with full input and fixed in training time, and thus will not be affected by the deviated rationales. The mechanisms of our model are supported by theoretical analysis and verified with empirical results.
Extensive experiments also demonstrate the great competitiveness of our method as compared to state-of-the-art methods.
In summary, our contribution lies in the following four aspects:
\begin{itemize}
    \item Problem identification: To the best of our knowledge, we are the first to identify the rationale shift problem in self-explaining rationalization, which may open up new directions for future research in this field.
    \item New framework: We propose a novel rationalization framework named DAR, which leverages a pretrained model to align the selected rationale to the full input with a discriminative alignment. 
    \item Theoretical insights: We theoretically show the importance of the alignment between the rationale and the full input (Lemma~\ref{lem:x get lower entropy than z}), the reason why the rationale shift problem can occur (Lemma~\ref{lem:conditioning increase divergence}), and the predictor of DAR can generalize to the full input (Theorem~\ref{theorem:x get high acc}), thus escaping rationale shift.
    \item Empirical results: Extensive experiments are conducted on two real-world datasets and  and two synthetic settings. Results on real-world datasets show that DAR improves the F1 score by up to $8.6\%$ as compared to the state-of-the-art method. 
\end{itemize}

The rest of the paper is organized as follows. Section~\ref{sec:related work} summarizes the related works. The problem definition of rationalization is given in Section~\ref{sec:problem definition}. The proposed method, motivation, and theoretical analysis are specified in Section~\ref{sec:proposed methods}. We show the experimental results on various datasets in Section~\ref{sec:experiments}. Finally, we conclude this study in Section~\ref{sec:conclusion}.
In order to streamline the main text for better readability, we have moved some non-critical experimental details and technical proofs to the appendix section.

\section{Related Work}\label{sec:related work}
In this article, we categorize the research in the field of \emph{eXplainable Artificial Intelligence} (XAI) into three main types: self-explaining, post-hoc, and generative explanations. Each of these will be detailed in the following subsections. Our primary focus will be on the methods of rationalization in the domain of self-explaining explanations.

\subsection{Self-explaining Rationalization} 
\cite{emnlp/LeiBJ16} first proposes the cooperative framework of rationalization named RNP. A unique advantage of rationalization is the certification of exclusion, which means any unselected input is guaranteed to have no contribution to prediction \cite{interlocking}. Based on this cooperative framework, many variants have been proposed to improve RNP from different aspects. 

  \textbf{Sampling Methods for Rationalization}. One series of research efforts focus on refining the sampling  process in the generator from different perspectives. For example, \cite{2018rationalegumble} utilized
Gumbel-softmax to smooth the binarized selection with parameters. \cite{hardkuma} replaced the Bernoulli sampling
distributions with rectified Kumaraswamy distributions. \cite{informationbottle} considered there is a trade-off between the prediction performance and the rationale length and imposed a discrete bottleneck objective to balance them. \cite{jain2020faith} exploited a saliency threshold to split the training regimes of the generator and predictor. Other methods like data augmentation with pretrained models \cite{counter}, training with human-annotated rationales \cite{Unirex}, intervention for causal rationales \cite{interventional} have also been tried. These methods are orthogonal to our research. 

  \textbf{Degeneration}. Another series of efforts focus on solving the degeneration problem of rationalization \cite{rethinking,dmr,interlocking}. These methods mainly exploit extra modules to provide calibrated information for the rationalization models. 
For instance, 3PLAYER \cite{rethinking} considers that the unselected text contains valuable information which is necessary to keep the semantics of the rationales unchanged, and thus employs a supplementary predictor to squeeze this information into the rationale. While the rationale selected by this method may contain the necessary information in the full text, they cannot exclude the noise from the rationale and thus there still exists the rationale shift problem. 
Similar to 3PLAYER, a recent method DARE \cite{dare} also takes the unselected parts into consideration. It tries to minimize the mutual information between the selected parts and unselected parts.
DMR \cite{dmr} feeds the full text and selected rationales to different predictors separately and then aligns their outputs. 
A2R \cite{interlocking} endows the predictor with the calibrated information of full text by introducing a soft rationale. Specifically, they use a supplementary predictor to process the masked text with soft attention as the input. To convey the formation of full text to the predictor, they minimize the JS-divergence between the output of the original predictor and of the supplementary predictor. Although DMR and A2R can fix the degeneration problem to some extend, the selected rationales may still deviate from the full text because the two predictors are fully or partially isolated from each other, and aligning their outputs does not necessarily aligns their inputs.
 
These methods are most related to our work which both provide calibrated information to the rationalization models. Nevertheless, they mainly focus on the special case of the rationale shift problem and train extra modules from scratch. These modules themselves may also be affected to some extent and thus have difficulties in fully solving the general case of rationale shift problem, because the deviation may be hard to be exactly captured by the affected modules.

\subsection{Post-hoc Explanation} 
Post-hoc methods for explainability have also been widely explored \cite{lime,cooperative,scott}. Among them, PGExplainer \cite{pgexplainer} and GDM \cite{GDM} are the methods most relevant to the line of rationalization. PGExplainer constructs a parametric explainer for Graph Neural Networks, and GDM generates hard masks as explanations for image classification. The difference is that they generate an explanation for an already-trained predictor. One shortcoming is that post-hoc methods usually get low predictive performance when using the explanation as the input. The reason can also be seen as a shift problem, i.e., the predictor is trained in the full input domain but tested in the explanation domain. Another shortcoming is that post-hoc explanation usually provides less transparency \cite{lipton2016mythos} and faithfulness \cite{rethinking}.
Our proposed method DAR can be seen as a combination of the self-explaining and post-hoc methods.

\subsection{Generative Explanation with Large Language Models}
Generative explanation is a research line that is close but orthogonal to our research. With the great success of large language models (LLMs), a new research line for XAI is chain-of-thought. By generating (in contrast to selecting) intermediate
reasoning steps before inferring the answer, the reasoning steps can be seen as a kind of explanation. The intriguing
technique is called chain-of-thought (CoT) reasoning \cite{cot}. However, LLMs sometimes exhibit unpredictable failure modes \cite{causalllm} or hallucination reasoning \cite{surveyofllm}, making this kind of generative explanation not trustworthy enough in some high-stakes scenarios. Also, some recent research finds that LLMs are not good at extractive tasks \cite{chatgptgeneral,evaluatingchatgpt,comprehensivechatgpt}.

\section{Problem Definition}\label{sec:problem definition}
For the sake of brevity and better benchmarks of comparison with other mainstream methods in the field of rationalization, we consider the input $X$ as text data in this study. 

\textbf{Notation}. 
In the following sections, $f_G(\cdot)$ and $f_P(\cdot)$ represent the generator and predictor. $\theta_G$ and $\theta_P$ represent the parameters of the generator and predictor. 
We consider the classification problem, where  ${X}$=$[x_1,x_2,\cdots,x_l]$ is the input text with $x_i$ being the $i-$th token and $l$ being the number of tokens. 
The label of ${X}$ is a one-hot vector $Y\in\{0,1\}^c$, where $c$ is the number of categories. $\mathcal{D}$ represents the dataset, which can be seen as a collection of samples drawn from the true data distribution $P(X,Y)$.

\textbf{Self-explaining rationalization}. The typical rationalization framework RNP consists of a generator and a predictor. The goal of the generator is to find a subset of the original input that best represents it. For each sample $(X,Y)\sim \mathcal{D}$, the generator firstly outputs a sequence of binary mask $M=[m_1,\cdots,m_l]\in \{0,1\}^l$ (in practice, it's binarized through gumbel-softmax). Then, it forms the rationale $Z$ by the element-wise product of $X$ and $M$:
\begin{equation}\label{eqa:getrat}
    Z=M\odot X=[m_1x_1,\cdots,m_lx_l].
\end{equation}
To simplify the notation, we denote the rationale $Z$ by $f_G(X)$, i.e., $Z=f_G(X)$. 
In self-explaining rationalization, the informativeness of the rationale $Z$ provided by the generator is measured by the negative cross entropy $-H_c(Y,\hat{Y}|Z)$, where $\hat{Y}$ is the output of the predictor with the input being $Z$. Consequently, the generator and the predictor are usually optimized cooperatively:
\begin{equation}\label{eqa:objpg}
\begin{split}
    &\mathop{\min}\limits_{\theta_G,\theta_P}H_c(Y,\hat{Y}|f_G(X)),\\
   & \ s.t. \ \hat{Y}=f_P(f_G(X)), \ (X,Y) \sim \mathcal{D}.
\end{split}
\end{equation}

\textbf{Short and coherent regularizer}. 
To make the selected rationales human-intelligible, the original RNP constrains the rationales by short and coherent regularization terms. In this paper, we use the most widely used constraints updated by \cite{car}:
\begin{equation}\label{eqa:regular}
\Omega (M) = \lambda_1 \lvert \frac{||M||_1}{l}-\alpha\rvert +\lambda_2\sum_t|m_t-m_{t-1}|, 
\end{equation} where $l$ denotes the number of tokens in the input text. The first term encourages that the percentage of the tokens being selected as rationales is close to a pre-defined level $\alpha$. The second term encourages the rationales to be coherent.

\section{Methodology and Theoretical Analysis}\label{sec:proposed methods}
In this section, we elaborate on the motivation for the proposed method and then specify the method DAR. After that, we theoretically analysis the properties of the method.

\subsection{Motivation}\label{sec:motivation}
\textbf{Rationale quality is positively related to the prediction accuracy on the full input.} To generate high-quality rationales, RNP generally adopts the prediction accuracy of the predictor over the rationales as the metric for rationale quality. Although this metric is simple and easy-understanding, more and more recent works find that the predictor can still achieve high prediction accuracy when the quality of rationales are low \cite{rethinking,dmr,interlocking}. A new metric for calibration is necessary for rationalization. Considering that the plausible rationales are semantically consistent with the input full text, a direct intuition is that a good predictor should also achieve high prediction accuracy over the full text when the quality of rationales is good and vice versa (please refer to Lemma~\ref{lem:x get lower entropy than z}).

To verify this intuition, we conduct an experiment where RNP achieves five different converged models with different sets of hyper-parameters to show the relationship between the prediction accuracy of the predictor on the full text and the rationale quality. The experimental results on \emph{Hotel-Service} dataset are shown in Fig.~\ref{fig:rationale_acc1}. More results and details are in Appendix~\ref{app: set up for rationale acc}. The orange line represents the prediction accuracy of the predictor using the full text as the input and the purple line represents the rationale quality measured by the overlap (F1 score) between the model-selected tokens and human-annotated rationales. 
It can be clearly concluded that the full-text prediction accuracy of the predictor is positively related to the rationale quality, which verifies intuition. Besides, the results indicate that, to improve the rationale quality of the rationalization, a predictor that has high prediction accuracy over the input full text will bring some benefits.

\begin{figure*}[t]
\quad \quad \quad \quad 
      \subfloat[Relationship]{
            \includegraphics[width=0.7\columnwidth]{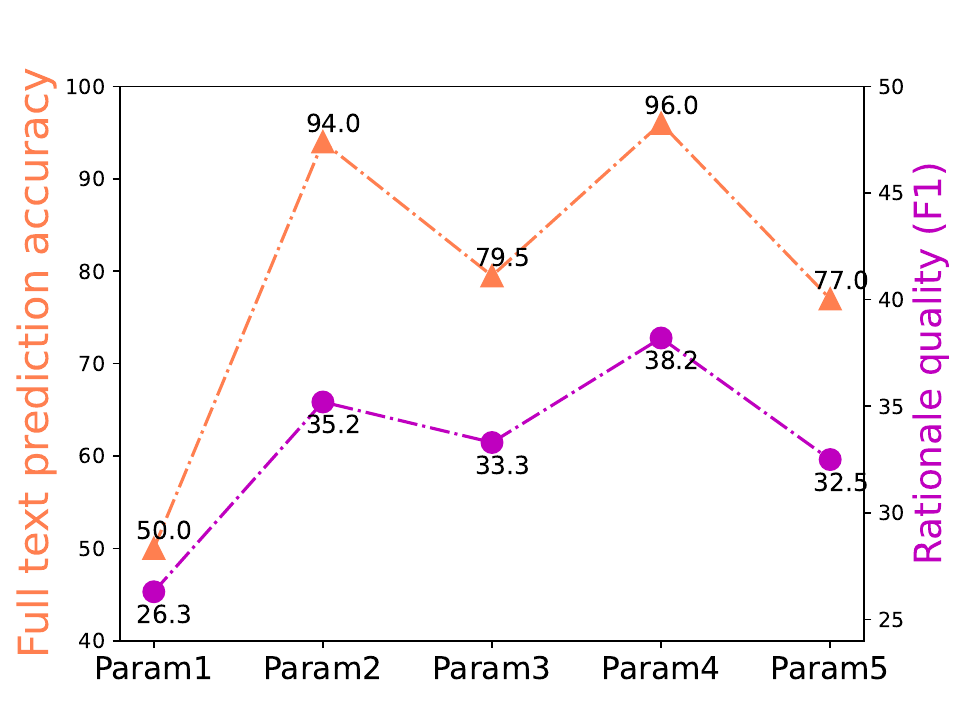}
        \label{fig:rationale_acc1}
        }
\hfil 
\quad \quad \quad
        \subfloat[Accuracy difference]{
            \includegraphics[width=0.7\columnwidth]{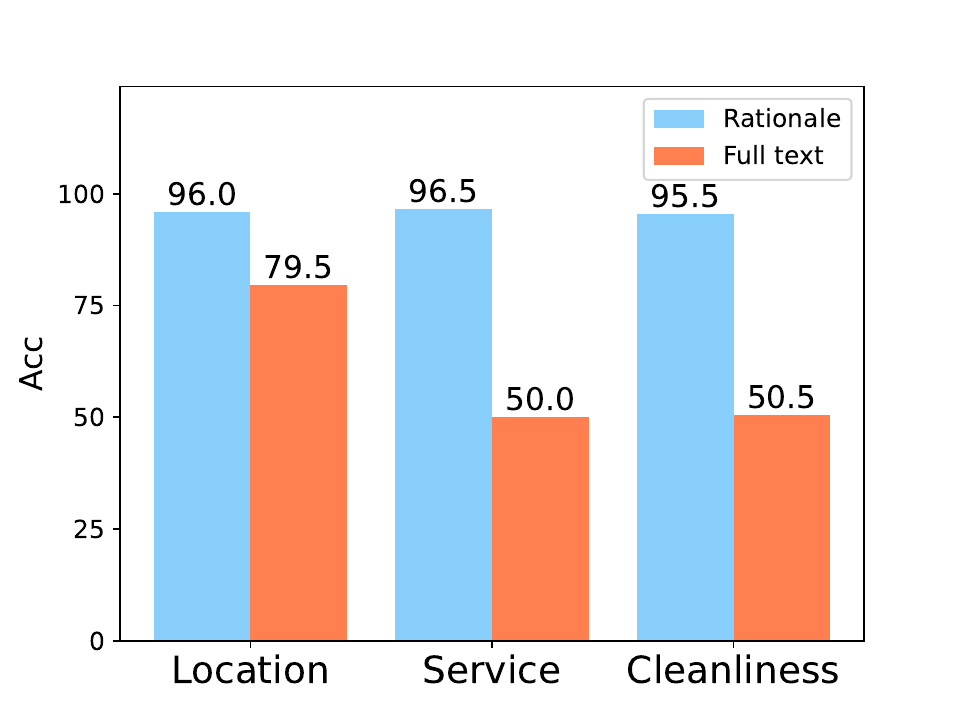}
            \label{fig:rationale_acc2}
        }

    \caption{Rationale shift of rationalization. The dataset is a multi-aspect binary classification dataset named HotelReview, which contains three aspects: location, service, and cleanliness. Each aspect is trained independently. (a) Relationship between the accuracy on full text and the rationale quality in \emph{service} aspect. (b) Accuracy difference of RNP on full text and on selected rationales.}
    \label{fig:rationale_acc}
\end{figure*}

\textbf{The rationale shift problem in RNP.}
It has been shown that the predictor tends to achieve high accuracy over the input full text when the semantics of the rationales are consistent with the input full text, i.e., high quality of rationales. Now, we seek further to investigate the consistency between the rationales and the input full text in RNP by diving into the prediction accuracy of its predictor over the full texts and the rationales, respectively. Specifically, we trained RNP on a multi-aspect binary classification dataset and present the two types of accuracy in Fig.~\ref{fig:rationale_acc2}. More experimental details can be found in Appendix~\ref{app: set up for rationale acc}. It can be seen that sometimes the predictor of RNP can not make classifications at all for the full text in the \emph{Service} aspect and \emph{Cleanliness} aspect, even though it predicts well with the rationale input. The results indicate that the predictor of RNP may learn rationales that heavily deviate from the input full text.

\begin{table}[t]
    \centering
        \caption{Predictive precision (\emph{P}), recall (\emph{R}), and F1 score (\emph{F1}) of RNP's predictor. }
    \resizebox{0.99\columnwidth}{!}{
    \begin{tabular}{|c | c c |c|c | c c |c |c | c c |c| }
\hline
  \multicolumn{4}{|c}{Location} & \multicolumn{4}{|c}{Service} & \multicolumn{4}{|c|}{Cleanliness}\\
\hline
S & P & R &\multicolumn{1}{c|}{F1} &S & P & R &\multicolumn{1}{c|}{F1} &S& P & R &\multicolumn{1}{c|}{F1}\\
\hline
9.0& 92.0&66.4 & 77.1&11.6 &100&1.0&2.0 &10.8 & nan & 0.0  &nan\\
\hline
\multicolumn{12}{l}{The results correspond  to the orange bar in Fig.~\ref{fig:rationale_acc}(b). $``$nan$"$: not a number.}\\
\multicolumn{12}{l}{\emph{S} indicates the average percentage of selected rationales. }\\
\multicolumn{12}{l}{ }
\end{tabular}
}

    \label{tab: sentence acc}
\end{table}

To further demystify whether the rationales that the predictor of RNP overfits include deviation, we present the detailed results of the predictor of RNP. Specifically, we show the detailed predictive precision and recall rate of the experiment in Table~\ref{tab: sentence acc} (Note that  the F1 score here is different from the one used in Fig.~\ref{fig:rationale_acc1} which indicates the overlap between the selected tokens and the human-annotated rationale.). As can be seen, for the \emph{Cleanliness} aspect, the predictor predicates all the full texts as negative by observing the precision and recall, and there is a similar case for the \emph{Service} aspect. Therefore, the predictor of RNP may overfit some unique biases delivered by the rationales, of which the phenomenon we denote as \emph{rationale shift}. 

To solve this problem, an intuitive idea is to improve the accuracy of the predictor on the full text such that the consistency between the rationale and the input full text can be implicitly achieved. In this paper, we take a further step by following this idea, where we employ a predictor pretrained with the full text and discriminatively align the rationales to the input full text with this predictor.

\begin{figure}[t]
    \centering
    \includegraphics[width=0.8\columnwidth]{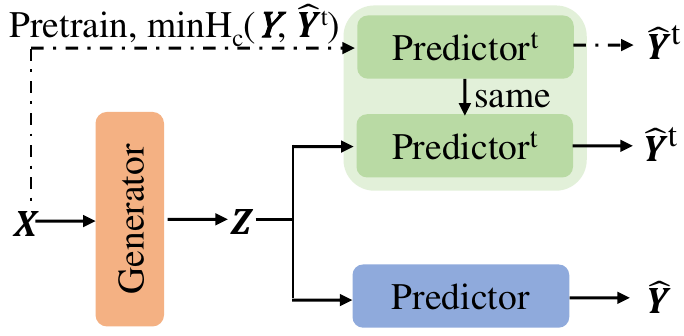}
    \caption{Our proposed DAR. The orange and blue modules are the same as RNP. The green module is the discriminative predictor.}
    \label{fig:gr}
\end{figure}

\subsection{The proposed Method}
Considering the limits of original rationalization, we propose a new method named DAR of which the architecture is shown in Fig.~\ref{fig:gr}. Compared to the vanilla RNP, DAR has an auxiliary \emph{predictor$^t$} (denoted as $f_{P_t}$) which is pretrained on the original full text and fixed in the training time of the rationalization models. 

Specifically, we first pretrain \emph{predictor$^t$} with 
\begin{equation}\label{eqa:best teacher}
\begin{split}
&\theta_{P_t}^*=\mathop{\arg\min}_{\theta_{P_t}}H_c(Y,\hat{Y}^t|X), \\
&s.t. \ \hat{Y}^t=f_{P_t}(X), \ (X,Y) \sim \mathcal{D}
\end{split}
\end{equation}
where $\theta_{P_t}$ represent the parameters of \emph{predictor$^t$} and $\theta_{P_t}^*$ is the optimal solution. Then, we fix the pretrained model $\theta_{P_t}^*$ and employ it as a discriminator to use its discriminative information to update rationalization models. The discriminative loss function is
\begin{equation}\label{eqa:regular for distribution}
\begin{split}
    &\mathop{\min}_{\theta_{G}}H_c(Y,\hat{Y}^t|f_G(X)), \\
    & \ s.t. \ \hat{Y}^t=f_P(f_G(X)),\  (X,Y) \sim \mathcal{D}
\end{split}
\end{equation}
Finally, joining (\ref{eqa:objpg}), (\ref{eqa:regular}), and (\ref{eqa:regular for distribution}) together, the overall loss function of the DAR is:
\begin{align}\label{eqa:objpgDPR}
    \begin{split}
        &\mathop{\min}\limits_{\theta_G,\theta_P}[H_c(Y,\hat{Y}|f_G(X))+H_c(Y,\hat{Y}^t|f_G(X)) + \Omega (M)], \\
        & \ s.t.\  \hat{Y}=f_P(f_G(X)), \ \hat{Y}^t=f_{P_t}(f_G(X)), \ (X,Y) \sim \mathcal{D}.
    \end{split}
\end{align}

\begin{figure}
    \centering
    \includegraphics[width=0.5\linewidth]{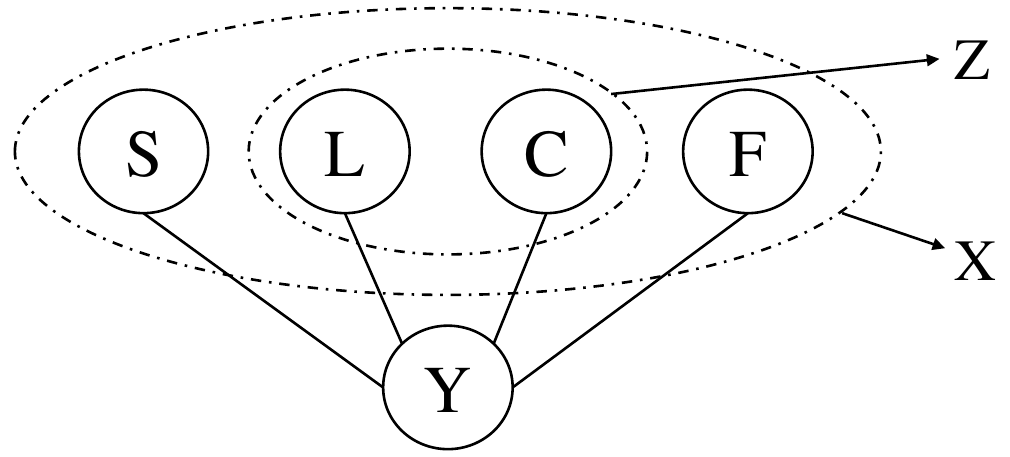}
    \caption{$X$ is a set of random variables, $Z$ is a subset of $X$.}
    \label{fig:relation between x and z}
\end{figure}

\subsection{Theoretical Analysis}

We first demonstrate why a good predictor needs to make the right prediction with the full text. Subsequently, we illustrate why the rationale shift problem can occur in the vanilla RNP framework. Finally, we show how the discriminative \emph{predictor$^t$} helps the original predictor to achieve it.

\subsubsection{Why is a good predictor expected to have better predictive performance
with the full input?}

\textbf{$X$ is more indicative to $Y$ than $Z$}. To begin with, we first show a toy example with Fig.~\ref{fig:relation between x and z}. $X$ is a set of random variables, or a random vector (not limited to text, text is a specific kind of vector), and $Z$ is part of this set. $Y$ is the overall quality of a hotel. $X$ is the review text. Each review discusses four aspects: \textbf{S}ervice, \textbf{L}ocation, \textbf{C}leanliness, \textbf{F}ood, and each aspect can be seen as a random variable. $Z$ consists of a subset (some dimensions) of $X$, e.g. (L, C). Generally, $X$ can be denoted as $X=(Z,X_{-Z})$, where $X_{-Z}$ is the remaining subset excluding $Z$.

We then easily get the mutual information between $Y$ and the full input is higher than that between $Y$ and the rationale candidate:
\begin{equation}
\begin{aligned}
    I(Y;X)=I(Y;Z,X_{-Z})&=I(Y;Z)+I(Y;X_{-Z}|Z)\\
    &\geq I(Y;Z),
\end{aligned}
\end{equation}
where the second equality is the chain rule for mutual information and $``\geq"$ comes from the property that mutual information is no smaller than zero.
Then, we have 
\begin{equation}
\begin{aligned}
      &I(Y;X)=H(Y)-H(Y|X)\\
      &\geq I(Y;Z)=H(Y)-H(Y|Z),
\end{aligned}
\end{equation}
that is to say $H(Y|X)\leq H(Y|Z)$.

\begin{lemma}\label{lem:x get lower entropy than z}
The entropy of $Y$ given the full input $X$ is lower than the entropy of $Y$ given the rationale candidate $Z$:
\begin{equation}
    H(Y|X)\leq H(Y|Z).
\end{equation}
\end{lemma}

\emph{\textbf{Remark}}. The practical prediction errors can be empirically measured by the cross-entropy (denoted with subscript $c$):
\begin{equation}
\begin{aligned}
    &H_c(Y,\hat{Y}|Z)=H(Y,Z)+D_{KL}(P_{Y|Z}||P_{\hat{Y}|Z}),\\
    &H_c(Y,\hat{Y}|X)=H(Y,X)+D_{KL}(P_{Y|X}||P_{\hat{Y}|X}).
\end{aligned}
\end{equation}
They consist of two parts. 
The entropies $H(Y,Z)$ and $H(Y,X)$ are the ideal (i.e., lower bound) prediction errors, and the KL-divergences are the predictor-dependent approximation errors. We have $H(Y|X)\leq H(Y|Z)$, which means the full input reduces more uncertainty and is more indicative to the label $Y$. Therefore, an \textbf{idealized} (akin to a human expert) predictor that can approximate the true distribution well is expected to have better prediction performance with the full input $X$. 

\subsubsection{Why can the rationale shift problem occur?}
\textbf{It is easier for a predictor to fit $Z$ rather than $X$}. Although $X$ is more indicative to $Y$, a predictor usually tends to overfit a distribution that contains fewer conditions. Consider the following lemma:
\begin{lemma}\label{lem:conditioning increase divergence}
    Given three random variables $Y, \hat{Y}, A$, where $Y$ and $\hat{Y}$ share the same label set $\mathcal{Y}$, it can be stated that conditioning on $A$ never leads to a reduction in divergence:
\begin{equation}
    D_{KL}(P_{Y|A}||P_{\hat{Y}|A})\geq D_{KL}(P_{Y}||P_{\hat{Y}}).
\end{equation}
\end{lemma}
The proof is in Appendix~\ref{proof:conditioning increase divergenc}. 

 \emph{\textbf{Remark}}. Here $A$ can stand for $X$ and $Z$ in rationalization, and $\hat{Y}$ can be seen as the predictor's output. And we then have 
\begin{equation}\label{eqa:z lower divergence}
\begin{aligned}
    &D_{KL}(P_{Y|X}||P_{\hat{Y}|X})\\
    &=D_{KL}(P_{Y|Z,X_{-Z}}||P_{\hat{Y}|Z,X_{-Z}})\geq D_{KL}(P_{Y|Z}||P_{\hat{Y}|Z}).
\end{aligned}
\end{equation}
 When we use $A$ to predict $Y$, the accuracy depends on two terms: the uncertainty of $P(Y|A)$ and how well the predictor approximates $P(Y|A)$ with $P(\hat{Y}|A)$, which is reflected by the cross-entropy: $H_c(Y,\hat{Y}|A)=H(Y|A)+ D_{KL}(P_{Y|A}||P_{\hat{Y}|A})$. Equation (\ref{eqa:z lower divergence}) indicates that, compared to overfitting trivial patterns in $Z$, it takes more effort for the predictor to fit the true semantics that are consistent between $Z$ and $X$. Thus we need more effort to enable the predictor to generalize to the full texts and to make $X$ be as indicative to $Y$ as expected.

\subsubsection{How can the proposed DAR align $Z$ to $X$?}
Then, we show how DAR can align the rationales and the full text, and thus enable the predictor to generalize to the full texts by training it with the aligned rationales.

\begin{lemma}\label{lem:best teacher}
   The cross entropy $H_c(Y,\hat{Y}^t|X)$ is lower-bounded by the entropy $H(Y|X)$, i.e., $H_c(Y,\hat{Y}^t|X)\geq H(Y|X)$, and the equality holds if and only if ${P}(Y|X)={P}(\hat{Y}^t|X)$.   
\end{lemma}

Lemma~\ref{lem:best teacher} is a basic idea of information theory and the proof is in Appendix~\ref{proof:best teacher}. By training the auxiliary \emph{predictor$^t$} to get the optimal solution of (\ref{eqa:best teacher}), we indirectly approximate the distribution ${P}(Y|X)$ with ${P}(\hat{Y}^t|X)$. Similarly, with sending the rationale $Z$ to \emph{predictor$^t$} and optimizing (\ref{eqa:regular for distribution}), we finally get ${P}(\hat{Y}^t|Z)={P}(Y|X)$. And for RNP's original loss function (\ref{eqa:objpg}), the optimal target is ${P}(\hat{Y}|Z)={P}(Y|X)$.
   Then, we have the conclusion:
\begin{theorem}\label{theorem:x get high acc}
    If the equations (\ref{eqa:best teacher}), (\ref{eqa:regular for distribution}), and (\ref{eqa:objpg}) all get the optimal solutions, we will have that the predictor will get the same output for the input being $X$ and $Z=f_G(X)$:
    \begin{equation}
        {P}(\hat{Y}|Z)={P}(Y|X)={P}(\hat{Y}|X).
    \end{equation}

\end{theorem}
The proof is in Appendix~\ref{proof:x get high acc}. Theorem~\ref{theorem:x get high acc} indicates that \emph{predictor} of DAR can also make the right prediction when the input is the full text. 

\begin{figure}[t]

    \centering

    \includegraphics[width=0.98\columnwidth]{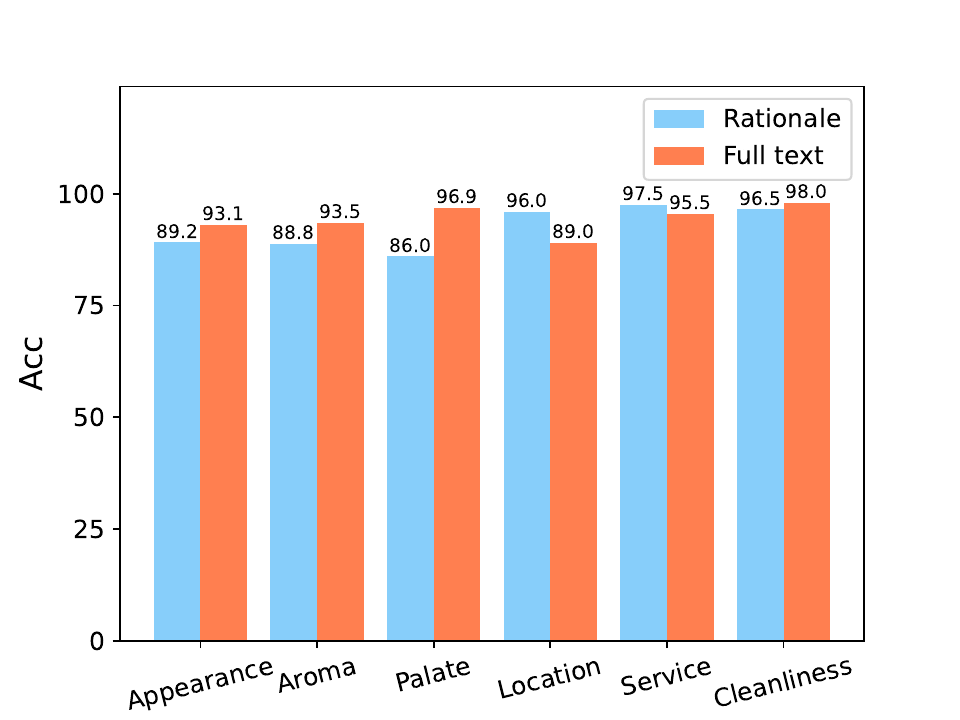}
    \caption{The prediction accuracy of our DAR with the inputs being the selected rationale and the full text respectively.}
    \label{fig:dpr rationale acc}

\end{figure}

\textbf{Empirical verification}.
To verify the practical performance, we show the predictor's accuracy of our DAR on two multi-aspect classification dataset in Fig.~\ref{fig:dpr rationale acc}. The blue bars represent the input is the selected rationale, and the orange bars represent the input is the full text. Each aspect is trained independently. Although the predictor of our DAR has never seen the full text during training, it can perform well with the full text input in most cases.

\begin{table*}[t]
    \centering
        \caption{Results on \emph{BeerAdvocate}. Each aspect is trained independently. The second best F1-scores are \underline{underlined}.}
\resizebox{1.995\columnwidth}{!}{
    \begin{tabular}{ |c c |c c| c c c|c c| c c c |c c| c c c| }
\hline
\multicolumn{2}{|c|}{\multirow{2}{*}{Methods}} & \multicolumn{5}{c|}{Appearance} & \multicolumn{5}{c|}{Aroma} & \multicolumn{5}{c|}{Palate}\\
\cline{3-17}
\multicolumn{2}{|c|}{} &S& Acc & P & R &\multicolumn{1}{c|}{F1} &S& Acc & P & R &\multicolumn{1}{c|}{F1} &S& Acc& P & R &\multicolumn{1}{c|}{F1}\\
\hline
\multicolumn{2}{|c|}{RNP$^*$\cite{emnlp/LeiBJ16}} &os& 85.7 & 83.9 & 71.2 & 72.8 &os&84.2&73.6 &67.9& 65.9 & os & 83.8 & 55.5 & 54.3 & 51.0\\
\multicolumn{2}{|c|}{re-DMR\cite{dmr}} &18.2&N/A& 71.1 &70.2 & 70.7 &15.4&N/A&59.8 & 58.9 & 59.3 &11.9&N/A&53.2 &50.9 & 52.0\\
\multicolumn{2}{|c|}{re-Inter\_RAT\cite{interventional}} & 18.2 & 89.9 & 57.5 &57.1 & 57.3 & 14.8 & 90.0 & 65.5 & 62.6&64.0&13.1&85.8&49.3&51.8&50.5\\
\multicolumn{2}{|c|}{re-A2R\cite{interlocking}} & 18.4 & 83.9 & 72.7 &72.3 & 72.5 & 15.4 & 86.3 & 63.6 & 62.9&63.2&12.4&81.2&57.4&57.3&57.4\\
\multicolumn{2}{|c|}{A2R$^*$\cite{interlocking}} & os & 86.3 & \textbf{84.7} &71.2 & \underline{72.9} & os & 84.9 & \textbf{79.3} & 71.3&\underline{70.0}&os&84.0&{64.2}&60.9&\underline{58.0}\\
\multicolumn{2}{|c|}{DAR(ours)} &18.6&89.2&{79.1}&\textbf{80.5}&\textbf{79.8
}&15.4&88.8&74.9&\textbf{73.9}&\textbf{74.4}&12.9&86.0&\textbf{64.8}&\textbf{68.6}&\textbf{66.6}
  \\\hline
\multicolumn{17}{l}{$``*"$: results obtained from the paper of A2R. $``$re-$"$: our reimplemented methods. $``$os$"$: one sentence. }
\end{tabular}
}

    \label{tab:beer}
\end{table*}

\begin{table*}[t]
    \centering
    \caption{Results on \emph{HotelReview}. Each aspect is trained independently. }
    \resizebox{1.99\columnwidth}{!}{
    \begin{tabular}{|c c c |c c| c c c|c c| c c c |c c| c c c| }
\hline
\multicolumn{3}{|c|}{\multirow{2}{*}{Methods}} & \multicolumn{5}{c|}{Location} & \multicolumn{5}{c|}{Service} & \multicolumn{5}{c|}{Cleanliness}\\
\cline{4-18}
\multicolumn{3}{|c|}{} &S& Acc & P & R &\multicolumn{1}{c|}{F1} &S& Acc & P & R &\multicolumn{1}{c|}{F1} &S& Acc& P & R &\multicolumn{1}{c|}{F1}\\
\hline
\multicolumn{3}{|c|}{RNP$^*$\cite{emnlp/LeiBJ16}} &10.9&-& 43.3 & 55.5 & 48.6  &11.0&-&40.0 &38.2& 39.1 & 10.6 & - & 30.5 & 36.0 & 33.0\\
\multicolumn{3}{|c|}{CAR$^*$\cite{car}} &10.6&N/A& 46.6 &58.1 & 51.7 &11.7&N/A&40.7 & 41.4 & 41.1 &9.9&N/A&32.3&35.7 & 33.9\\
\multicolumn{3}{|c|}{DMR$^{\dagger}$\cite{dmr}} & 10.7 & N/A & 47.5 &60.1 & \underline{53.1} &11.6&N/A & 43.0 & 43.6 & \underline{43.3}&10.3&N/A&31.4&36.4&33.7\\
\multicolumn{3}{|c|}{re-Inter\_RAT\cite{interventional}} &  11.0 &95.5 &34.7 &44.8 & 39.1 &12.5 & 98.5 & 35.4 & 39.1&37.2&9.6&97.0&33.4&36.7&\underline{34.9}\\
\multicolumn{3}{|c|}{re-A2R\cite{interlocking}} &  8.5 &87.5 &43.1 &43.2 & 43.1 &11.4 & 96.5 & 37.3 & 37.2&37.2&8.9&94.5&{33.2}&33.3&33.3\\
\multicolumn{3}{|c|}{DAR(ours)} &9.8&97.0&\textbf{52.3}&\textbf{60.4}&\textbf{56.0}&11.4&{97.5}&\textbf{48.6}&\textbf{48.3}&\textbf{48.4}&10.8&{97.0}&\textbf{35.7}&\textbf{44.2}&\textbf{39.5}
  \\\hline
  \multicolumn{18}{l}{$``*"$: results obtained from the paper of CAR, $\dagger$: results obtained from the paper of DMR. $``$re-$"$: our reimplemented method. }
\end{tabular}
}

    \label{tab:hotel}

\end{table*}

\begin{table}[t]
    \caption{The complexity
of different models. }
    \centering
    \resizebox{0.99\columnwidth}{!}{
    \setlength\tabcolsep{3pt}
    \begin{tabular}{|c|c c c c c|}
    \hline
     & RNP\cite{emnlp/LeiBJ16} & CAR\cite{car}& DMR\cite{dmr} & A2R\cite{interlocking}& DAR(Ours)  \\
     \hline
  modules &1gen$+$1pred&1gen$+$2pred&1gen$+$3pred&1gen$+$2pred&1gen$+$2pred \\     parameters & $2\times$ &$3\times$ &$4\times$ &$3\times$&$3\times$\\
  \hline
  \multicolumn{6}{l}{$``$gen$"$: generator. $``$pred$"$: predictor. }
    \end{tabular}
    }

    \label{tab:para}
\end{table}

\section{Experiments}\label{sec:experiments}
In this section, we evaluate our method DAR\footnote{Code: \url{https://github.com/jugechengzi/dar}} in various settings to demonstrate its effectiveness. 

\subsection{Experimental Setup}

 \textbf{Baselines}.  We compare our DAR with the original rationalization method RNP and several recent published models that achieve state-of-the-art results on real-world benchmarks, including DMR \cite{dmr}, A2R \cite{interlocking}, and Inter\_RAT \cite{interventional}, all of which have been discussed in detail in Section~\ref{sec:related work}. Table~\ref{tab:para} is a comparison between the complexity of different models.
We also compare with some methods that are re-implemented by \cite{danqi} with powerful BERT generator and predictor, including VIB \cite{informationbottle} and SPECTRA \cite{spectral}, as a supplement. 

\textbf{Models}. Following DMR, A2R and Inter\_RAT, we use 200-dimension bi-directional gated recurrent units (GRUs) \cite{gru} followed by one linear layer for each of the players, and the word embedding is 100-dimension Glove \cite{glove}. A recent research \cite{danqi} finds that it is still a challenging task to finetune large pretrained models on the RNP framework and most recent methods perform very bad with over-parameterized pretrained models (see Table 4 in \cite{danqi} and Appendix~\ref{app:bert}), so we use GRUs to keep the same settings as the baselines in the main experiments for a fair comparison, but we also conduct experiments with BERT-base-uncased \cite{bert} as a supplement. The optimizer is Adam \cite{adam}. The reparameterization trick for binarized sampling is Gumbel-softmax \cite{2016gumble,2018rationalegumble}, which is also the same as most of the baselines.

\begin{table*}[t]

\centering
\caption{Results of methods with low sparsity on \emph{BeerAdvocate}.  }
\resizebox{1.99\columnwidth}{!}{
\begin{tabular}{|c c c |c | c c c|c | c c c | c| c c c| }
\hline
\multicolumn{3}{|c|}{\multirow{2}{*}{Methods}} & \multicolumn{4}{|c|}{Appearance} & \multicolumn{4}{c|}{Aroma} & \multicolumn{4}{c|}{Plate}\\
\cline{4-15}
\multicolumn{3}{|c|}{} &S & P & R &\multicolumn{1}{c|}{F1} &S & P & R &\multicolumn{1}{c|}{F1} &S& P & R &\multicolumn{1}{c|}{F1}\\
\hline

\multicolumn{3}{|c|}{RNP$^*$} & 11.9& 72.0 & 46.1 &56.2 &10.7  & 70.5 & 48.3&\underline{57.3}&10.0&53.1&42.8&47.5\\
\multicolumn{3}{|c|}{CAR$^*$} & 11.9  & 76.2 & 49.3 & 59.9 & 10.3  & 50.3 &33.3&40.1&10.2&56.6&46.2&50.9\\
\multicolumn{3}{|c|}{DMR$^{\dagger}$} & 11.7  &{83.6} &52.8 & \underline{64.7} &11.7 & 63.1 & 47.6 & 54.3&10.7&55.8&48.1&\underline{51.7}\\
\multicolumn{3}{|c|}{DAR(ours)} &11.5 & \textbf{90.7}&\textbf{59.3}&\textbf{71.7}&10.2& \textbf{86.4}&\textbf{56.7}&\textbf{68.5}&10.8&\textbf{62.2}&\textbf{54.7}&\textbf{58.2}
  \\\hline
  \multicolumn{15}{l}{\tiny $``*"$: results obtained from the paper of CAR. $\dagger$: results obtained from DMR. }
 \end{tabular}
}

\label{tab:ls}

\end{table*}

\begin{table}[t]

\centering
\caption{Results with BERT encoder on \emph{Beer-Appearance} dataset. }
\resizebox{0.95\columnwidth}{!}{
\begin{tabular}{|c c c |c c| c c |c |}
\hline
\multicolumn{3}{|c|}{\multirow{2}{*}{Methods}} & \multicolumn{5}{c|}{Beer-Appearance} \\
\cline{4-8}
\multicolumn{3}{|c|}{} &S& Acc & P & R &\multicolumn{1}{c|}{F1} \\
\hline

\multicolumn{3}{|c|}{VIB$^*$\cite{informationbottle}} & -&-& - & - &20.5 \\
\multicolumn{3}{|c|}{SPECTRA$^*$\cite{spectral}} & -&-&- & - &\underline{28.6} \\
\multicolumn{3}{|c|}{CR$^{\dagger}$\cite{cr}} & -&-&- & - &27.4 \\
\multicolumn{3}{|c|}{re-RNP} & 15.8&92.1& 22.1 & 19.1 &20.5 \\
\multicolumn{3}{|c|}{DAR(ours)} &18.7&89.9 & \textbf{72.1}&\textbf{73.6}&\textbf{72.8}
  \\\hline
  \multicolumn{8}{l}{\scriptsize $``*"$: results obtained from \cite{danqi}.}\\
  \multicolumn{8}{l}{ \scriptsize $\dagger$: CR is a method try to find causal rationales.}
 \end{tabular}
}

\label{tab:bert}
\end{table}

\textbf{Datasets}.
Following DMR, we consider two widely used multi-aspect classification datasets for rationalization tasks. Note that each of them contains three distinct aspects, which are trained independently in our experiments. Consequently, these two datasets can be considered as six distinct datasets to some extent.
\textbf{BeerAdvocate} \cite{beer} is a multi-aspect sentiment prediction dataset on reviewing beers. There is a high correlation among the rating scores of different aspects in the same review, making it difficult to directly learn a rationalization model from the original data. We use the subsets decorrelated by \cite{emnlp/LeiBJ16} as most previous works did \cite{car,dmr,interlocking,dare}.
\textbf{HotelReview} \cite{hotel} is another multi-aspect sentiment prediction dataset on reviewing hotels. The dataset contains reviews of hotels from three aspects including location, cleanliness, and service. Each review has a rating on a scale of 0-5 stars. We follow the previous works \cite{dmr,interlocking} to  
binarize the labels. Both datasets contain human-annotated rationales on the test set only. We preprocess both datasets in the same way as \cite{dmr} for a fair comparison and the details are in Appendix~\ref{app:datasets}.

 \textbf{Implementation Details}.
Note that the rationales of BeerAdvocate are annotated on a sentence level, so A2R does sentence-level selection on it. In this paper, we do a more flexible token level selection on both datasets as most previous methods do \cite{rethinking,car,dmr}, which is a harder process. We also re-implement A2R based on its source codes to do the token-level selection. The sparsity of selected rationales is set to be similar to the percentage of human-annotated rationales. 

The hyper-parameters of reimplemented models are manually tuned multiple times to get the best results. For our DAR, the early stopping technique is conducted according to the predictive accuracy of the development set. For our reimplemented baselines, although we have tried our best to tune the hyper-parameters, chances are that the hyper-parameters are still not the best, and to compensate for this potential issue, we choose their best results when they get the best $F1$ score on the test set. More details are in Appendix~\ref{app: implementation details}.

 \textbf{Metrics}. Following the baselines \cite{dmr,interlocking,interventional,dare}, we mainly focus on the rationale quality, which is measured by the overlap between the model-selected tokens and human-annotated tokens. $P,R,F1$ indicate the precision, recall, and F1 score, respectively. $S$ indicates the average sparsity of the rationales, i.e., the percentage of selected tokens to the whole text. $Acc$ indicates the predictive accuracy with the selected rationale being the input. Note that DMR and CAR \cite{car} adopt adversarial games and use the label as part of the input to select texts, so they don't contain the prediction accuracy.

\subsection{Evaluation on Standard Benchmarks}

\begin{table*}[t]
 \caption{Results of skewed predictor that induces rationale shift on \emph{BeerAdvocate}. }  
\footnotesize
    \centering
    \resizebox{1.99\columnwidth}{!}{
    \begin{tabular}{|c |c |c| c c |c|c| c c |c|c| c c |c|}
    \hline
    \multirow{2}{*}{Aspect} &\multirow{2}{*}{Setting}& \multicolumn{4}{c|}{RNP*} & \multicolumn{4}{c|}{A2R*}& \multicolumn{4}{c|}{DAR(ours)} \\
\cline{3-14}
{}&{} &Acc&P & R &F1&Acc & P & R &F1&Acc&P & R &F1\\
\hline
\multicolumn{1}{|c|}{{\multirow{3}{*}{Aroma}}} &\multicolumn{1}{c|}{skew10} &82.6&68.5 &63.7 & 61.5 &84.5&\textbf{78.3}&70.6&69.2& 89.2 &75.9&\textbf{72.0}&\textbf{73.9}\\
\multicolumn{1}{|c|}{}&\multicolumn{1}{c|}{skew15} &80.4&54.5& 51.6&49.3 &81.8&58.1&53.3&51.7& 89.4 &\textbf{76.5}&\textbf{72.1}&\textbf{74.2}\\
\multicolumn{1}{|c|}{}&\multicolumn{1}{c|}{skew20} &76.8&10.8 & 14.1 &11.0 &80.0&51.7&47.9&46.3&88.8&\textbf{74.6}&\textbf{73.9}&\textbf{74.2}\\
\cline{1-14}
\multicolumn{1}{|c|}{{\multirow{3}{*}{Palate}}} &\multicolumn{1}{c|}{skew10} &77.3&5.6 &7.4 & 5.5 &82.8&50.3&48.0&45.5 &81.7&\textbf{56.9}&\textbf{63.4}&\textbf{60.0}\\
\multicolumn{1}{|c|}{}&\multicolumn{1}{c|}{skew15} &77.1&1.2 & 2.5 & 1.3 &80.9&30.2&29.9&27.7&83.0&\textbf{56.9}&\textbf{63.6}&\textbf{60.1}\\
\multicolumn{1}{|c|}{}&\multicolumn{1}{c|}{skew20} &75.6&0.4 & 1.4 & 0.6 &76.7&0.4&1.6&0.6& 82.6 &\textbf{55.5}&\textbf{64.9}&\textbf{59.8}\\
\hline
\multicolumn{14}{l}{$``*"$: results obtained from A2R.}
    \end{tabular}
    }

     \label{tab:predskew}
\end{table*}

\begin{table*}[h]
    \caption{Results of skewed generator that induces rationale shift in the \emph{Palate} aspect of \emph{BeerAdvocate}.}
\footnotesize
    \centering
    \resizebox{1.99\columnwidth}{!}{
    \begin{tabular}{|c |c c c| c c| c| c c c| c c| c|}
    \hline
    \multicolumn{1}{|c|}{\multirow{2}{*}{Setting}} & \multicolumn{6}{c|}{RNP} & \multicolumn{6}{c|}{DAR(ours)} \\
\cline{2-13}
\multicolumn{1}{|c|}{} &Pre\_acc&S&Acc & R & R &F1&Pre\_acc&S&Acc & P & R &F1\\
\hline
\multicolumn{1}{|c|}{skew60.0} &63.1&13.1&87.2 &42.8 & 45.1 &43.9&63.2&12.9&83.1&\textbf{54.7}& \textbf{56.7} &\textbf{55.7}\\
\multicolumn{1}{|c|}{skew65.0} &66.6&14.0&83.9 &40.3 &45.4&42.7&67.9&12.8&82.9&\textbf{52.7}&\textbf{54.4}&\textbf{53.6}\\
\multicolumn{1}{|c|}{skew70.0} &71.3&14.7&84.1&10.0 & 11.7 & 10.8&70.2&13.4&79.1&\textbf{49.4}& \textbf{53.1} &\textbf{51.2}\\
\multicolumn{1}{|c|}{skew75.0} &75.5 & 14.7 &87.6 &8.1&9.6&8.8&75.5&14.1&78.6&\textbf{46.9}& \textbf{52.9} &\textbf{49.7}\\
\hline
    \end{tabular}
    }

    \label{tab:skew generator}

\end{table*}

\textbf{Comparison with baselines on two datasets}. Table~\ref{tab:beer} and~\ref{tab:hotel} show the results of selected rationales on two datasets. All the methods choose a similar percentage of tokens that is close to the human-annotated sparsity by adjusting the sparsity regularization term in (\ref{eqa:regular}). Obviously, we beat all the baselines in all six aspects. 
In particular, the F1 score of our method DAR is $66.6\%$ in the \emph{Beer-Palate} dataset, which outperforms the state-of-the-art result by up to $8.6\%$, i.e., $58.0\%$ of A2R. And for the \emph{HotelReview} dataset, the best improvement is $5.1\%$ (Service aspect). 

\textbf{Comparison with baselines in low rationale sparsity}. To show the robustness of our DAR, we also conduct an experiment where the sparsity of selected rationales is very low, which is the same as CAR \cite{car} and DMR \cite{dmr}. 
The results are shown in Table~\ref{tab:ls}. The results of RNP and CAR are obtained from \cite{car}, and the results of DMR are obtained from \cite{dmr}. We still achieve great improvements over other methods. In this situation, we  outperform the state-of-the-art result by up to $11.2\%$ (\emph{Aroma} aspect).

\textbf{Experiments with BERT encoder}. 
Most recent rationalization models fail to select the true rationale when using large pretrained models like BERT \cite{bert} as the encoders of the generator and the predictor. For example, experiments in \cite{danqi} show that two improved rationalization models VIB \cite{informationbottle} and SPECTRA \cite{spectral} perform much worse when using BERT encoders. One reason may be that the powerful large pretrained models can recognize very small deviations, causing the damage of rationale shift more serious. For a fair comparison, we do not use BERT in our main experiments. To show that our method is suitable for BERT, we further conduct an experiment where the GRUs is replaced with BERT-base-uncased as a supplement. Following \cite{danqi}, we use the \emph{Beer-Appearance} dataset. The results are shown in Table~\ref{tab:bert}. Our DAR gets significant improvements as compared to the two widely used rationalization methods VIB and SPECTRA.

\subsection{Evaluation on Synthetic Settings}
To better show the influence of rationale shift, we further conduct two synthetic experiments.

\textbf{Skewed predictor} is a synthetic setting designed by A2R \cite{interlocking}. It produces a kind of rationale shift by deliberately inducing the predictor to overfit to manually added rationale deviation, which creates an extreme training obstacle named interlocking.  
We first train the predictor separately using only the first sentence of each input text, and further cooperatively train the predictor initialized with the pretrained parameters and the generator randomly initialized using normal input.
In \emph{BeerAdvocate}, the first sentence is usually about appearance. So, the predictor will overfit to the aspect of \emph{Appearance}, which is uninformative for \emph{Aroma} and \emph{Palate}. “\emph{skew}$k$” denotes the predictor is pre-trained for $k$ epochs. To make a fair comparison, we keep the pre-training process the same as that of A2R: we use a batch size of 500 and a learning rate of 0.001. 

The results are shown in Table~\ref{tab:predskew}. The results of RNP and A2R are obtained from \cite{interlocking}. For all the settings, we outperform both RNP and A2R. Especially, for the relatively easy task on \emph{Aroma}, the performance of DAR is hardly affected by this poor initialization as compared to the results in Table~\ref{tab:beer}. And for the relatively hard task on \emph{Palate}, it is only slightly affected, while RNP and A2R can hardly work. It can be concluded that DAR is much more robust than the other two methods in this situation.

\textbf{Skewed generator}. As opposed to skewed predictor, we also conduct an experiment where we give the generator special initialization to induce it to select rationales with a great shift. We pretrain the generator separately using the text classification label as the mask label of the first token. In other words, for texts of class 1, we force the generator to select the first token, and for texts of class 0, we force the generator not to select the first token. So, the generator learns the category implicitly by whether the first token is chosen and the predictor only needs to learn this position information to make a right prediction.

We use \emph{Beer-Palate} because A2R shows that Palate is harder than other aspects. $k$ in “skew$k$” denotes the threshold of the skew: we pretrain the generator as a special classifier of the first token for a few epochs until its predictive accuracy is higher than $k$. Since the accuracy increases rapidly in the first a few epochs, obtaining a model that precisely achieves the pre-defined accuracy is almost impossible. So, we use “$Pre\_acc$” to denote the actual predictive accuracy of the generator-classifier when the pre-training process stops. Higher “$Pre\_acc$” means more severe shift.

In this case, RNP fails to find the human-annotated rationales but our DAR still works well.

\section{Conclusion and future work}\label{sec:conclusion}
In this paper, we try to develop transparent and trustworthy deep learning models through self-explaining rationalization. We first identify rationalization suffers from an important problem named rationale shift, which stems from the algorithmic bias of the vanilla cooperative game. 
To address this problem, we propose a novel method named DAR which employs a model pretrained with the full input to discriminatively align the rationale and the full input. We demonstrate the benefits of aligning the rationales and the full text with both empirical observations (Fig.~\ref{fig:rationale_acc}) and theoretical analysis (Lemma~\ref{lem:x get lower entropy than z}), and further prove that the pretrained model facilitates their consistency. Extensive experiments show that our method greatly improves the rationale quality as compared to traditional methods. Besides, how to utilize pretrained language models in the self-explaining framework of rationalization is a big challenge \cite{danqi}, but our approach is applicable to pretrained models and brings some progress to this line of research. 

Given the flexibility of the self-explaining rationalization framework, our proposed methods hold great potential for application to diverse fields such as computer vision and graph learning. Therefore, a direction for future research is to explore the extension of our techniques beyond the scope of text-based data.

\section{Limitations}
One limitation may be that the obstacles in utilizing powerful pretrained language models under the self-explaining rationalization framework remain mysterious. While we have made some progress in this direction, the empirical results with pretrained models do not exhibit significant improvements compared to those achieved with GRUs. Further endeavors should be undertaken in the future to explore what happened to these models. However, it is somewhat out of the scope of this paper, and we leave it as future work.

\section*{Acknowledgment}
We appreciate Mcauley et al. \cite{beer} and Wang et al. \cite{hotel} for their datasets. We also appreciate Chang et al. \cite{car} and Huang et al. \cite{dmr} for their released code.

\appendix

\section{Experimental Setup}

\begin{table}[t]
   \centering
       \caption{Statistics of datasets used in this paper}
   \setlength\tabcolsep{2pt}
    \begin{tabular}{|c l| c c| c c| c c c|}
    \hline
         \multicolumn{2}{|c|}{\multirow{2}{*}{Datasets}}&\multicolumn{2}{c|}{Train}&\multicolumn{2}{c|}{Dev}&\multicolumn{3}{c|}{Annotation}  \\
         \multicolumn{2}{|c|}{}& Pos&Neg&Pos&Neg&Pos&Neg&S\\
         \hline
        \multirow{3}{*}{Beer}&Appearance&16891&16891 &6628&2103&923&13&18.5\\
        {}&Aroma&15169&15169&6579&2218&848&29&15.6\\
        {}&Palate&13652&13652&6740&2000&785&20&12.4\\
        \hline
        \multirow{3}{*}{Hotel}&Location&7236&7236 &906&906&104&96&8.5\\
        {}&Service&50742&50742&6344&6344&101&99&11.5\\
        {}&Cleanliness&75049&75049&9382&9382&99&101&8.9\\
        \hline
    \end{tabular}

    \label{tab:dataset}
\end{table}

\subsection{Datasets}\label{app:datasets}
\textbf{BeerAdvacate}. Following \cite{car,dmr,interlocking}, we consider a classification setting by treating reviews with ratings $\leq$ 0.4 as
negative and $\geq$ 0.6 as positive. Then we randomly select examples from the original training set to
construct a balanced set.

\textbf{Hotel Reviews}.
Similar to BeerAdvacate, we treat reviews with ratings $<$ 3 as
negative and $>$ 3 as positive.

More details are in Table~\ref{tab:dataset}. $Pos$ and $Neg$ denote the number of positive and negative examples in each set. $Sparsity$ denotes the average percentage of tokens in human-annotated rationales to the whole texts.

\begin{table}[t]
    \centering
        \caption{Hyper-parameters of experiments in Fig.~\ref{fig:rationale_acc1}}
    \begin{tabular}{|c|c|c|c|}
\hline
        & lr ($10^{-4}$) & batchsize & hidden\_dim \\
        \hline
        Param1& 1 & 256 & 100      \\
        Param2& 1 & 256 & 200 \\
         Param3& 2 & 256 & 200       \\
        Param4& 1 & 512 & 200        \\
        Param5& 2 & 512 & 200        \\
            
\hline
    \end{tabular}

    \label{tab:hyperparameter for rationale acc}
\end{table}

\subsection{Implementation details}\label{app: implementation details}

For BeerAdvocate, DMR reported results with rationales of very low sparsity. So, we also re-implement DMR with its source codes and adjust the sparsity (i.e., $\alpha$ in (\ref{eqa:regular})) to get the sparsity similar to human-annotated rationales. The random seed is kept the same across all the experiments rather than manually selected. The hyper-parameters of reimplemented models are manually tuned multiple times under this fixed random seed to get the best results (we recognize that the best hyper-parameters of DMR vary a lot across different random seeds). The early stop is conducted according to the predictive accuracy on the development set. For our method DAR, we take the results when the model get best prediction accuracy on the development set. For our reimplemented baselines, although we have tried our best to tune the hyper-parameters, chances are that the hyper-parameters are not the best, and to compensate for this potential issue, we choose their best results when they get the best $F1$ score on the test set.

\begin{figure}[t]
    \centering
    \includegraphics[width=0.7\columnwidth]{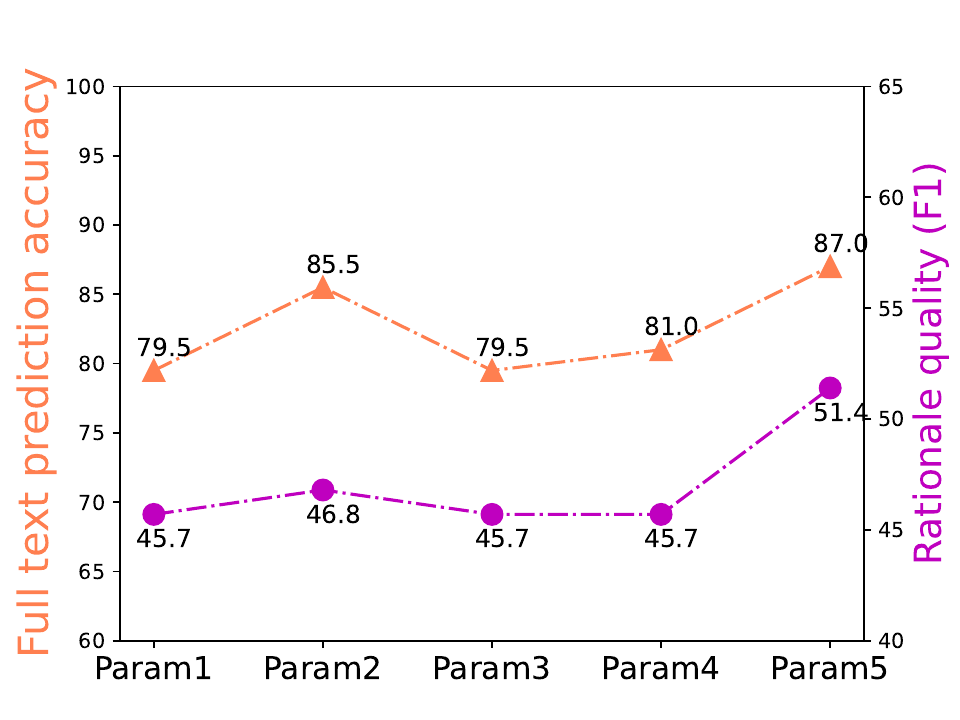}
    \caption{Relationship between
the accuracy on full text and the rationale quality on \emph{Hotel-Location} dataset.}
    \label{fig:rationale acc as0}
\end{figure}

\subsection{Set up of the experiment in Fig.~\ref{fig:rationale_acc}}\label{app: set up for rationale acc}

\begin{figure}[t]
    \centering
    \includegraphics[width=0.7\columnwidth]{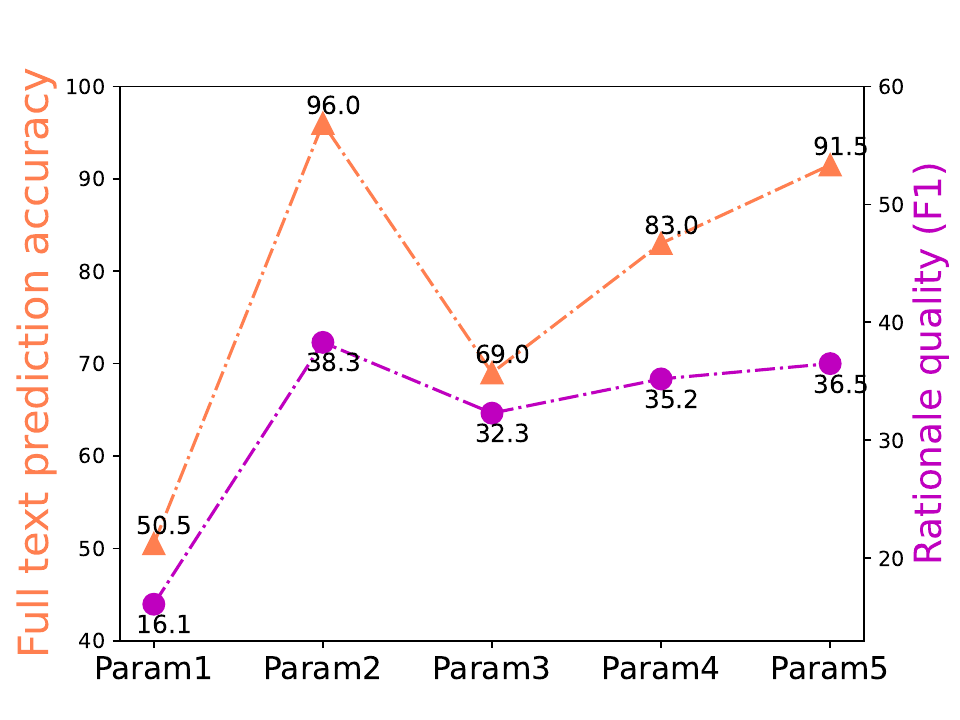}
    \caption{Relationship between
the accuracy on full text and the rationale quality on \emph{Hotel-Cleanliness} dataset.}
    \label{fig:rationale acc as2}
\end{figure}

We train the vanilla RNP with five sets of hyper-parameters for each aspect of \emph{HotelReview}, which are shown in Table~\ref{tab:hyperparameter for rationale acc}. Fig.~\ref{fig:rationale_acc2} shows the results with \emph{Param1} for all three aspects. Fig.~\ref{fig:rationale_acc1} shows the comparison between the results of the five sets of hyper-parameters in the \emph{Service} aspect. The results in the \emph{Location} and \emph{Cleanliness} aspect are in Fig.~\ref{fig:rationale acc as0} and Fig.~\ref{fig:rationale acc as2}, respectively.

  \begin{table}[t]
  \caption{Results (F1 score) with BERT.}
   \centering
   \setlength\tabcolsep{2pt}
    \begin{tabular}{|c |c| c|}
    \hline
    Methods & Beer-Appearance &Hotel-Cleanliness\\
    \hline
    VIB \cite{informationbottle}&20.5&23.5\\
    SPECTRA \cite{spec}&28.6&19.5\\
    \hline
    \multicolumn{3}{l}{\footnotesize The results are from Table 4 of \cite{danqi}.}
         
    \end{tabular}

    \label{tab:res_bert}
\end{table}

  \begin{table}[t]
  \caption{Results (F1-score) with BERT. }
   \centering
   \setlength\tabcolsep{2pt}
    \begin{tabular}{|c|c |c| c|}
    \hline
    Methods & Beer-Appearance &Beer-Aroma&Beer-Palate\\
    \hline
    VIB\cite{informationbottle}&21.5&30.1&19.8\\
    CR\cite{cr}&27.4&39.0&22.6\\
    \hline
    \multicolumn{4}{l}{The results are from Table 1 of \cite{cr}.}
         
    \end{tabular}

    \label{tab:res_bert_cr}
\end{table}

\subsection{Discussion on BERT encoder}\label{app:bert}
 In the field of rationalization, researchers generally focus on frameworks of the models and the methodology. Methods most related to our work do not use Bert or other pre-trained encoders \cite{car,invarant,dmr,rethinking,interlocking}. We use GRUs and GloVe to ensure the same experimental setup as our baselines for a fair comparison.

 More importantly, how to finetune large models on the rationalization framework is still a significant challenge. Some recent studies \cite{danqi} show that the methods with BERT encoders perform much worse than those with simple GRUs on BeerAdvocate and HotelReviews, which is shown in Table~\ref{tab:res_bert}. VIB and SPECTRA are two RNP-based model. When using BERT, these two methods perform much worse than the vanilla RNP with GRUs. And the experiments in a recent paper also support this idea (please refer to Table~\ref{tab:res_bert_cr}).

\section{Theorem Proofs}

\subsection{Proof of Lemma~\ref{lem:conditioning increase divergence}}\label{proof:conditioning increase divergenc}
\begin{equation}
\begin{aligned}
    &D_{KL}(P_{Y|A}||P_{\hat{Y}|A})-D_{KL}(P_{Y}||P_{\hat{Y}})\\
=&\sum_{a\in\mathcal{A}}\sum_{y\in \mathcal{Y}}P_{Y,A}(y,a)\cdot\ln\frac{P_{Y|A}(y|a)}{P_{\hat{Y}|A}(y|a)}-\sum_{y\in \mathcal{Y}}P_{Y}(y)\cdot\ln \frac{P_{Y}(y)}{P_{\hat{Y}}(y)} \\
=&\sum_{a\in\mathcal{A}}\sum_{y\in \mathcal{Y}}P_{Y,A}(y,a)\cdot\ln\frac{P_{Y|A}(y|a)}{P_{\hat{Y}|A}(y|a)}\\
&-\sum_{y\in \mathcal{Y}}(\sum_{a\in \mathcal{A}} P_{Y,A}(y,a))\cdot\ln \frac{P_{Y}(y)}{P_{\hat{Y}}(y)}\\
=&\sum_{a\in\mathcal{A}}\sum_{y\in \mathcal{Y}}P_{Y,A}(y,a)\cdot \ln \frac{P_{Y|A}(y|a)P_{\hat{Y}}(y)}{P_{\hat{Y}|A}(y|a)P_{Y}(y)}\\
\overset{\text{*}}{\geq}& \sum_{a\in\mathcal{A}}\sum_{y\in \mathcal{Y}}P_{Y,A}(y,a)(1-\frac{P_{\hat{Y}|A}(y|a)P_{Y}(y)}{P_{Y|A}(y|a)P_{\hat{Y}}(y)}) \quad \quad \quad \quad  \\
=& 1- \sum_{y\in \mathcal{Y}}(\frac{P_{Y}(y)}{P_{\hat{Y}}(y)}\sum_{a\in\mathcal{A}}P_A(a)P_{\hat{Y}|A}(y|a))\\
=&1-\sum_{y\in \mathcal{Y}}\frac{P_{Y}(y)}{P_{\hat{Y}}(y)}P_{\hat{Y}(y)}\\
=&0.
\end{aligned}
\end{equation}
$(*)$ is from $\ln x\geq 1-\frac{1}{x}$.

\subsection{Proof of Lemma~\ref{lem:best teacher}}\label{proof:best teacher}
It is a basic idea in information theory that the entropy of a distribution $P$ is upper bounded by the cross entropy of using $Q$ to approximate it.
For any two distribution $P$ and $Q$, we have 
\begin{equation}
\begin{aligned}  
    &H_c(P,Q)=H(P)+D_{KL}(P||Q),\\
    &D_{KL}(P||Q)\geq 0.
\end{aligned}
\end{equation}
So, we easily get $H_c(P,Q)\geq H(P)$. 
Similarly, we have 
\begin{equation}
\begin{aligned}
    H_c(Y\hat{Y}^t|X)&=H(Y|X)+D_{KL}(P(Y|X)||P(\hat{Y}^t|X))\\
    &\geq H(Y|X).
\end{aligned}
\end{equation}
The proof of Lemma~\ref{lem:best teacher} is completed.

\subsection{Proof of Theorem~\ref{theorem:x get high acc}}\label{proof:x get high acc}
According to Lemma~\ref{lem:best teacher}, we know when (\ref{eqa:best teacher}) gets the optimal solution, we have 
\begin{equation}
    {P}(Y|X)={P}(\hat{Y}^t|X).
\end{equation}
Similarly, when (\ref{eqa:regular for distribution}) gets the optimal solution, we have 
\begin{equation}\label{eqa:same yx}
    {P}(Y|X)={P}(\hat{Y}^t|Z=f_G(X)).
\end{equation}
So, we then have 
\begin{equation}\label{eqa:same yt}
    {P}(\hat{Y}^t|X)={P}(\hat{Y}^t|Z=f_G(X)).
\end{equation}
When (\ref{eqa:objpg}) gets the optimal solution, we have 
\begin{equation}\label{eqa:left dis}
    {P}(Y|X)={P}(\hat{Y}|Z=f_G(X)).
\end{equation}
Combining it with (\ref{eqa:same yx}) and (\ref{eqa:same yt}), we further have 
\begin{equation}\label{eqa:right dis}
    {P}(\hat{Y}|Z)={P}(\hat{Y}|X).
\end{equation}
Combining (\ref{eqa:left dis}) and (\ref{eqa:right dis}), the proof of Theorem~\ref{theorem:x get high acc} is completed.

\bibliographystyle{ieeetr}
\bibliography{custom}

\begin{thebibliography}{10}

\bibitem{lipton2016mythos}
Z.~C. Lipton, ``The mythos of model interpretability: In machine learning, the concept of interpretability is both important and slippery.,'' {\em Queue}, vol.~16, no.~3, pp.~31--57, 2018.

\bibitem{xiang2019interpretable}
L.~Xiang, H.~Zhang, H.~Ma, Y.~Zhang, J.~Ren, and Q.~Zhang, ``Interpretable complex-valued neural networks for privacy protection,'' in {\em International Conference on Learning Representations}, 2019.

\bibitem{miller2019explanation}
T.~Miller, ``Explanation in artificial intelligence: Insights from the social sciences,'' {\em Artificial intelligence}, vol.~267, pp.~1--38, 2019.

\bibitem{sun2021interpreting}
X.~Sun, D.~Yang, X.~Li, T.~Zhang, Y.~Meng, Q.~Han, G.~Wang, E.~Hovy, and J.~Li, ``Interpreting deep learning models in natural language processing: A review,'' {\em arXiv preprint arXiv:2110.10470}, 2021.

\bibitem{sigmod-debug}
R.~Pradhan, J.~Zhu, B.~Glavic, and B.~Salimi, ``Interpretable data-based explanations for fairness debugging,'' in {\em {SIGMOD} '22: International Conference on Management of Data, Philadelphia, PA, USA, June 12 - 17, 2022}, pp.~247--261, {ACM}, 2022.

\bibitem{li2022backdoor}
Y.~Li, Y.~Jiang, Z.~Li, and S.-T. Xia, ``Backdoor learning: A survey,'' {\em IEEE Transactions on Neural Networks and Learning Systems}, 2022.

\bibitem{danqi}
H.~Chen, J.~He, K.~Narasimhan, and D.~Chen, ``Can rationalization improve robustness?,'' in {\em Proceedings of the 2022 Conference of the North American Chapter of the Association for Computational Linguistics: Human Language Technologies, {NAACL} 2022, Seattle, WA, United States, July 10-15, 2022}, pp.~3792--3805, Association for Computational Linguistics, 2022.

\bibitem{Unirex}
A.~Chan, M.~Sanjabi, L.~Mathias, L.~Tan, S.~Nie, X.~Peng, X.~Ren, and H.~Firooz, ``{UNIREX:} {A} unified learning framework for language model rationale extraction,'' in {\em International Conference on Machine Learning, {ICML} 2022, 17-23 July 2022, Baltimore, Maryland, {USA}}, vol.~162 of {\em Proceedings of Machine Learning Research}, pp.~2867--2889, {PMLR}, 2022.

\bibitem{interlocking}
M.~Yu, Y.~Zhang, S.~Chang, and T.~S. Jaakkola, ``Understanding interlocking dynamics of cooperative rationalization,'' in {\em Advances in Neural Information Processing Systems 34: Annual Conference on Neural Information Processing Systems 2021, NeurIPS 2021, December 6-14, 2021, virtual}, pp.~12822--12835, 2021.

\bibitem{invarant}
S.~Chang, Y.~Zhang, M.~Yu, and T.~S. Jaakkola, ``Invariant rationalization,'' in {\em Proceedings of the 37th International Conference on Machine Learning, {ICML} 2020, 13-18 July 2020, Virtual Event}, vol.~119 of {\em Proceedings of Machine Learning Research}, pp.~1448--1458, {PMLR}, 2020.

\bibitem{emnlp/LeiBJ16}
T.~Lei, R.~Barzilay, and T.~S. Jaakkola, ``Rationalizing neural predictions,'' in {\em Proceedings of the 2016 Conference on Empirical Methods in Natural Language Processing, {EMNLP} 2016, Austin, Texas, USA, November 1-4, 2016}, pp.~107--117, The Association for Computational Linguistics, 2016.

\bibitem{dmr}
Y.~Huang, Y.~Chen, Y.~Du, and Z.~Yang, ``Distribution matching for rationalization,'' in {\em Thirty-Fifth {AAAI} Conference on Artificial Intelligence, {AAAI} 2021, Thirty-Third Conference on Innovative Applications of Artificial Intelligence, {IAAI} 2021, The Eleventh Symposium on Educational Advances in Artificial Intelligence, {EAAI} 2021, Virtual Event, February 2-9, 2021}, pp.~13090--13097, {AAAI} Press, 2021.

\bibitem{car}
S.~Chang, Y.~Zhang, M.~Yu, and T.~S. Jaakkola, ``A game theoretic approach to class-wise selective rationalization,'' in {\em Advances in Neural Information Processing Systems 32: Annual Conference on Neural Information Processing Systems 2019, NeurIPS 2019, December 8-14, 2019, Vancouver, BC, Canada}, pp.~10055--10065, 2019.

\bibitem{rethinking}
M.~Yu, S.~Chang, Y.~Zhang, and T.~S. Jaakkola, ``Rethinking cooperative rationalization: Introspective extraction and complement control,'' in {\em Proceedings of the 2019 Conference on Empirical Methods in Natural Language Processing and the 9th International Joint Conference on Natural Language Processing, {EMNLP-IJCNLP} 2019, Hong Kong, China, November 3-7, 2019}, pp.~4092--4101, Association for Computational Linguistics, 2019.

\bibitem{liufr}
W.~Liu, H.~Wang, J.~Wang, R.~Li, C.~Yue, and Y.~Zhang, ``{FR}: Folded rationalization with a unified encoder,'' in {\em Advances in Neural Information Processing Systems}, 2022.

\bibitem{GDM}
H.~Yuan, L.~Cai, X.~Hu, J.~Wang, and S.~Ji, ``Interpreting image classifiers by generating discrete masks,'' {\em {IEEE} Trans. Pattern Anal. Mach. Intell.}, vol.~44, no.~4, pp.~2019--2030, 2022.

\bibitem{pgexplainer}
D.~Luo, W.~Cheng, D.~Xu, W.~Yu, B.~Zong, H.~Chen, and X.~Zhang, ``Parameterized explainer for graph neural network,'' in {\em Advances in Neural Information Processing Systems 33: Annual Conference on Neural Information Processing Systems 2020, NeurIPS 2020, December 6-12, 2020, virtual}, 2020.

\bibitem{AAAI20sterberg}
H.~Zhang, W.~Chen, Z.~Huang, M.~Li, Y.~Yang, W.~Zhang, and J.~Wang, ``Bi-level actor-critic for multi-agent coordination,'' in {\em The Thirty-Fourth {AAAI} Conference on Artificial Intelligence, {AAAI} 2020, The Thirty-Second Innovative Applications of Artificial Intelligence Conference, {IAAI} 2020, The Tenth {AAAI} Symposium on Educational Advances in Artificial Intelligence, {EAAI} 2020, New York, NY, USA, February 7-12, 2020}, pp.~7325--7332, {AAAI} Press, 2020.

\bibitem{gan}
I.~J. Goodfellow, J.~Pouget{-}Abadie, M.~Mirza, B.~Xu, D.~Warde{-}Farley, S.~Ozair, A.~C. Courville, and Y.~Bengio, ``Generative adversarial nets,'' in {\em Advances in Neural Information Processing Systems 27: Annual Conference on Neural Information Processing Systems 2014, December 8-13 2014, Montreal, Quebec, Canada}, pp.~2672--2680, 2014.

\bibitem{2018rationalegumble}
Y.~Bao, S.~Chang, M.~Yu, and R.~Barzilay, ``Deriving machine attention from human rationales,'' in {\em Proceedings of the 2018 Conference on Empirical Methods in Natural Language Processing, Brussels, Belgium, October 31 - November 4, 2018}, pp.~1903--1913, Association for Computational Linguistics, 2018.

\bibitem{hardkuma}
J.~Bastings, W.~Aziz, and I.~Titov, ``Interpretable neural predictions with differentiable binary variables,'' in {\em Proceedings of the 57th Conference of the Association for Computational Linguistics, {ACL} 2019, Florence, Italy, July 28- August 2, 2019, Volume 1: Long Papers}, pp.~2963--2977, Association for Computational Linguistics, 2019.

\bibitem{informationbottle}
B.~Paranjape, M.~Joshi, J.~Thickstun, H.~Hajishirzi, and L.~Zettlemoyer, ``An information bottleneck approach for controlling conciseness in rationale extraction,'' in {\em Proceedings of the 2020 Conference on Empirical Methods in Natural Language Processing, {EMNLP} 2020, Online, November 16-20, 2020}, pp.~1938--1952, Association for Computational Linguistics, 2020.

\bibitem{jain2020faith}
S.~Jain, S.~Wiegreffe, Y.~Pinter, and B.~C. Wallace, ``Learning to faithfully rationalize by construction,'' in {\em Proceedings of the 58th Annual Meeting of the Association for Computational Linguistics, {ACL} 2020, Online, July 5-10, 2020}, pp.~4459--4473, Association for Computational Linguistics, 2020.

\bibitem{counter}
M.~Plyler, M.~Green, and M.~Chi, ``Making a (counterfactual) difference one rationale at a time,'' in {\em Advances in Neural Information Processing Systems 34: Annual Conference on Neural Information Processing Systems 2021, NeurIPS 2021, December 6-14, 2021, virtual}, pp.~28701--28713, 2021.

\bibitem{interventional}
L.~Yue, Q.~Liu, L.~Wang, Y.~An, Y.~Du, and Z.~Huang, ``Interventional rationalization,'' 2023.

\bibitem{dare}
L.~Yue, Q.~Liu, Y.~Du, Y.~An, L.~Wang, and E.~Chen, ``{DARE:} disentanglement-augmented rationale extraction,'' in {\em NeurIPS}, 2022.

\bibitem{lime}
M.~T. Ribeiro, S.~Singh, and C.~Guestrin, ``"why should {I} trust you?": Explaining the predictions of any classifier,'' in {\em Proceedings of the 22nd {ACM} {SIGKDD} International Conference on Knowledge Discovery and Data Mining, San Francisco, CA, USA, August 13-17, 2016}, pp.~1135--1144, {ACM}, 2016.

\bibitem{cooperative}
S.~Havrylov, G.~Kruszewski, and A.~Joulin, ``Cooperative learning of disjoint syntax and semantics,'' in {\em Proceedings of the 2019 Conference of the North American Chapter of the Association for Computational Linguistics: Human Language Technologies, {NAACL-HLT} 2019, Minneapolis, MN, USA, June 2-7, 2019, Volume 1 (Long and Short Papers)}, pp.~1118--1128, Association for Computational Linguistics, 2019.

\bibitem{scott}
P.~Fernandes, M.~Treviso, D.~Pruthi, A.~Martins, and G.~Neubig, ``Learning to scaffold: Optimizing model explanations for teaching,'' {\em Advances in Neural Information Processing Systems}, vol.~35, pp.~36108--36122, 2022.

\bibitem{cot}
J.~Wei, X.~Wang, D.~Schuurmans, M.~Bosma, B.~Ichter, F.~Xia, E.~H. Chi, Q.~V. Le, and D.~Zhou, ``Chain-of-thought prompting elicits reasoning in large language models,'' in {\em NeurIPS}, 2022.

\bibitem{causalllm}
E.~K{\i}c{\i}man, R.~Ness, A.~Sharma, and C.~Tan, ``Causal reasoning and large language models: Opening a new frontier for causality,'' {\em arXiv preprint arXiv:2305.00050}, 2023.

\bibitem{surveyofllm}
Z.~Ji, N.~Lee, R.~Frieske, T.~Yu, D.~Su, Y.~Xu, E.~Ishii, Y.~J. Bang, A.~Madotto, and P.~Fung, ``Survey of hallucination in natural language generation,'' {\em ACM Computing Surveys}, vol.~55, no.~12, pp.~1--38, 2023.

\bibitem{chatgptgeneral}
C.~Qin, A.~Zhang, Z.~Zhang, J.~Chen, M.~Yasunaga, and D.~Yang, ``Is chatgpt a general-purpose natural language processing task solver?,'' {\em arXiv preprint arXiv:2302.06476}, 2023.

\bibitem{evaluatingchatgpt}
B.~Li, G.~Fang, Y.~Yang, Q.~Wang, W.~Ye, W.~Zhao, and S.~Zhang, ``Evaluating chatgpt's information extraction capabilities: An assessment of performance, explainability, calibration, and faithfulness,'' {\em arXiv preprint arXiv:2304.11633}, 2023.

\bibitem{comprehensivechatgpt}
J.~Ye, X.~Chen, N.~Xu, C.~Zu, Z.~Shao, S.~Liu, Y.~Cui, Z.~Zhou, C.~Gong, Y.~Shen, {\em et~al.}, ``A comprehensive capability analysis of gpt-3 and gpt-3.5 series models,'' {\em arXiv preprint arXiv:2303.10420}, 2023.

\bibitem{spectral}
T.~Miyato, T.~Kataoka, M.~Koyama, and Y.~Yoshida, ``Spectral normalization for generative adversarial networks,'' in {\em 6th International Conference on Learning Representations, {ICLR} 2018, Vancouver, BC, Canada, April 30 - May 3, 2018, Conference Track Proceedings}, OpenReview.net, 2018.

\bibitem{gru}
K.~Cho, B.~van Merrienboer, {\c{C}}.~G{\"{u}}l{\c{c}}ehre, D.~Bahdanau, F.~Bougares, H.~Schwenk, and Y.~Bengio, ``Learning phrase representations using {RNN} encoder-decoder for statistical machine translation,'' in {\em Proceedings of the 2014 Conference on Empirical Methods in Natural Language Processing, {EMNLP} 2014, October 25-29, 2014, Doha, Qatar, {A} meeting of SIGDAT, a Special Interest Group of the {ACL}}, pp.~1724--1734, {ACL}, 2014.

\bibitem{glove}
J.~Pennington, R.~Socher, and C.~D. Manning, ``Glove: Global vectors for word representation,'' in {\em Proceedings of the 2014 Conference on Empirical Methods in Natural Language Processing, {EMNLP} 2014, October 25-29, 2014, Doha, Qatar, {A} meeting of SIGDAT, a Special Interest Group of the {ACL}}, pp.~1532--1543, {ACL}, 2014.

\bibitem{bert}
J.~Devlin, M.~Chang, K.~Lee, and K.~Toutanova, ``{BERT:} pre-training of deep bidirectional transformers for language understanding,'' in {\em Proceedings of the 2019 Conference of the North American Chapter of the Association for Computational Linguistics: Human Language Technologies, {NAACL-HLT} 2019, Minneapolis, MN, USA, June 2-7, 2019, Volume 1 (Long and Short Papers)}, pp.~4171--4186, Association for Computational Linguistics, 2019.

\bibitem{adam}
D.~P. Kingma and J.~Ba, ``Adam: {A} method for stochastic optimization,'' in {\em 3rd International Conference on Learning Representations, {ICLR} 2015, San Diego, CA, USA, May 7-9, 2015, Conference Track Proceedings}, 2015.

\bibitem{2016gumble}
E.~Jang, S.~Gu, and B.~Poole, ``Categorical reparameterization with gumbel-softmax,'' in {\em 5th International Conference on Learning Representations, {ICLR} 2017, Toulon, France, April 24-26, 2017, Conference Track Proceedings}, OpenReview.net, 2017.

\bibitem{cr}
W.~Zhang, T.~Wu, Y.~Wang, Y.~Cai, and H.~Cai, ``Towards trustworthy explanation: On causal rationalization,'' in {\em Proceedings of the 40th International Conference on Machine Learning (ICML'23)}, vol.~202 of {\em Proceedings of Machine Learning Research}, pp.~41715--41736, PMLR, 23--29 Jul 2023.

\bibitem{beer}
J.~J. McAuley, J.~Leskovec, and D.~Jurafsky, ``Learning attitudes and attributes from multi-aspect reviews,'' in {\em 12th {IEEE} International Conference on Data Mining, {ICDM} 2012, Brussels, Belgium, December 10-13, 2012}, pp.~1020--1025, {IEEE} Computer Society, 2012.

\bibitem{hotel}
H.~Wang, Y.~Lu, and C.~Zhai, ``Latent aspect rating analysis on review text data: a rating regression approach,'' in {\em Proceedings of the 16th {ACM} {SIGKDD} International Conference on Knowledge Discovery and Data Mining, Washington, DC, USA, July 25-28, 2010}, pp.~783--792, {ACM}, 2010.

\bibitem{spec}
N.~M. Guerreiro and A.~F.~T. Martins, ``{SPECTRA:} sparse structured text rationalization,'' in {\em Proceedings of the 2021 Conference on Empirical Methods in Natural Language Processing, {EMNLP} 2021, Virtual Event / Punta Cana, Dominican Republic, 7-11 November, 2021}, pp.~6534--6550, Association for Computational Linguistics, 2021.

\end{thebibliography}

\clearpage

\end{document}